\documentclass[sigconf]{acmart}

\AtBeginDocument{%
  \providecommand\BibTeX{{%
    \normalfont B\kern-0.5em{\scshape i\kern-0.25em b}\kern-0.8em\TeX}}}

\setcopyright{acmcopyright}
\copyrightyear{2018}
\acmYear{2018}
\acmDOI{10.1145/1122445.1122456}

\acmConference[Woodstock '18]{Woodstock '18: ACM International Conference of Human-Robot Interaction}{8 March 2021}{Online}
\acmBooktitle{Woodstock '18: ACM International Conference of Human-Robot Interaction,
8 March 2021, Online}
\acmPrice{15.00}
\acmISBN{978-1-4503-XXXX-X/18/06}

\begin{document}

\title{AIR4Children: Artificial Intelligence and Robotics for Children}



\author{Rocio Montenegro}
\affiliation{%
  \institution{AIR4Children}
  \city{Vancouver}
  \country{Canada}}

\author{Elva Corona}
\affiliation{%
  \institution{AIR4Children}
  \city{Xicothzinco}
  \country{M\'exico}}

\author{Donato Perez-Badillo}
\affiliation{%
  \institution{AIR4Children}
  \city{Xicohtzinco}
  \country{M\'exico}}

\author{Dago Cruz}
\affiliation{%
  \institution{AIR4Children}
  \city{San Diego California}
  \country{USA}}

\author{Miguel Xochicale}
\affiliation{%
  \institution{AIR4Children}
  \city{London}
  \country{UK}}
\email{air4children@gmail.com}

\renewcommand{\shortauthors}{AIR4Children}

\begin{abstract}
We introduce AIR4Children, Artificial Intelligence for Children, as a way to (a) tackle aspects for inclusion, accessibility, transparency, equity, fairness and participation and (b) to create affordable child-centred materials in AI and Robotics (AIR).
We present current challenges and opportunities for a child-centred approaches for AIR. 
Similarly, we touch on open-sourced software and hardware technologies to make a more inclusive, affordable and fair participation of children in areas of AIR. 
Then, we describe the avenues that AIR4Children can take with the development of open-sourced software and hardware based on our initial pilots and experiences.
Similarly, we propose to follow the philosophy of Montessori education to help children to not only develop computational thinking but also to internalise new concepts and learning skills through activities of movement and repetition.
Finally, we conclude with the opportunities of our work and mainly we pose the future work of putting in practice what is proposed here to evaluate the potential impact on AIR to children, instructors, parents and their community. 
\end{abstract}



\begin{teaserfigure}
  \includegraphics[width=\textwidth]{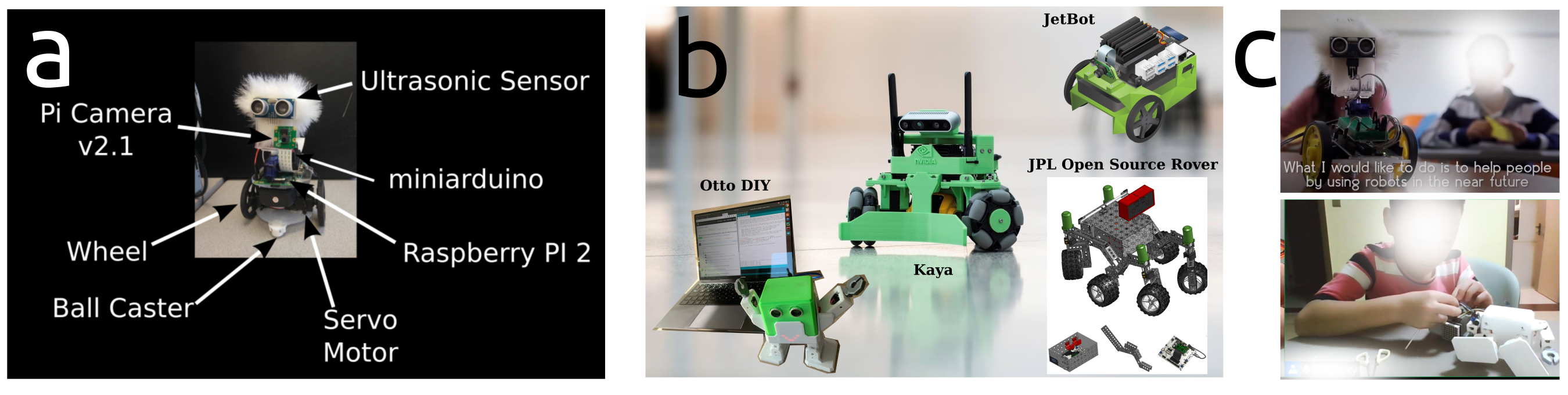}
  \caption{(a) Robot prototype (b) open-source robots for ai and robotics, (c) piloting teaching materials with children.}
  \label{fig:teaser}
\end{teaserfigure}

\maketitle

\section{Introduction} 
The scientific and technological progress in the fields of Artificial Intelligence and Robotics (AIR) has been rapidly moving forward over the past decade with special focus in countries such as United States, China and Europe where support of such fields is part of their agenda \cite{Savage2020}. 
Artificial Intelligence (AI), for instance, is changing and helping various aspects of our daily lives, from entertainment to complex medicine discovery tasks. 
Then the democratization of current AI technologies allows people without in-depth skills in maths, statistics, and programming to create their own AI solutions. Examples of these new open-source platforms with a high-level programming approach are PyTorch, TensorFlow, Keras, and Scikit-learn. 
However, most of these technologies use specific syntax and grammar and work with particular programming languages which makes the learning curve a complex task for children. 
Therefore, there is research to be done in order to design and create open-source platforms that bring children closer to AI.
Something similar happen in the case of Robotics where software-based AI solutions can deploy solutions on real robots (such as JIBO, NAO, others), however these robots are generally inaccessible for children worldwide, particularly for those of under-represented communities.
Such challenges of unreachable technologies in AI, and extrapolated to Robotics, to young audiences 
also include their affordability and open accessibility \cite{UNICEF2020}.

Therefore, this work coins the term of Artificial Intelligence and Robotics for Children (AIR4Children) as a way to make AI and Robotics available to young audiences from under-represented communities as well as to create and to design tools based on open-sourced projects.
Additionally, AIR4Children aims to create, to design and to distill curriculums based on non-traditional education (i.e. Montessori) for children of different socio-economical background, developmental stages and learning abilities. 

For this work, section 2 reviews open source projects as the basis of the material for AIR4Children.
Section 3 introduces AIR4Children as a open source project with open teaching materials based on Montessori education. 
We then conclude with current status of AIR for children, the creation of open source materials for hardware, software and teaching resources and mentions the future work for pilots and distillation of teaching materials.

\section{Open Source Software and Hardware for AI and Robotics}
Open source resources "refers to something people can modify and share because its design is publicly accessible" \cite{opensource2021}.
Such open source community was initiated in 1978 by Donald Knuth who designed \TeX, typesetting system, which is a role model for open source projects where organisational phases of its development and the relative and simple accessibility to users were crucial to its success \cite{gaudeul2007}.
Then, in 1983 Richard Stallman, with the frustration to not freely inspect, modify or share software, founded the GNU project to then create a GNU manifesto \cite{stallman1985}.
Both \TeX and GNU project were the corner stone of what is known today as the Open Software Initiative, founded by Bruce Perens and Eric S. Raymond in 1998, stating that projects must be free re distributable, code must be available and distributable, modification must be allowed, etc \cite{brasseur2018}.
Following a similar spirit that software can be used, studied, copied, modified, and redistributed without restriction, projects of open source hardware started to emerge in mid 2000s (e.g., OpenCores, RepRap (3D printing), Arduino, Adafruit and SparkFun) \cite{pearce2013}.

Then, in the last decade, another wave of scientific innovation for open source resources started with open source software frameworks (e.g. pytorch, tensorflow, etc.) along with places to distribute these (e.g. GitHub, gitlab, bitbucket, etc) \cite{matelabs2017}.
However, our understanding is that little has been done for child-centred AI and Robotics. 
For instance, Otto DIY is an educational open source robot founded in 2016 by Camilo Parra, where the community of Otto has more than 20,000 users from 20 countries and more than 100 re-designs of the robot \cite{OttoDIY:2016}.
Another example is the JPL Open Source Rover, created by engineers at NASA and initially released in April 2018, which it is designed with detailed instructions for constructions for mainly high school students, and open source technical specifications, 3D models and assembly instructions \cite{OSR:2018}.
Recently, engineers at NVIDIA in 2019, released nano JetBot platform  "to give the hands on experience needed to create entirely new AI projects" \cite{nanoJetBot:2019}. 

On other hand, there is an inherent engagement in children with robots as a way to use robots to enact education solutions \cite{druga2019}. 
They take advantage of real-time deployments from computers/mobile into real robots based on the Do-It-Yourself approach. 
Thus, the platforms encourage programming skills in children. 
Shybo is a robot that combines open-source hardware and software \cite{Lupetti2017}.
Shybo can perceive sounds and react through non-verbal behaviors (movements and lights). 
The creators claim that Shybo can also be used in educational contexts to support playful learning experiences. 
Sparky is an Arduino-based mobile robot. Creators provide schematics, 3D model files, and source code underneath are all open source \cite{sparky2012}. 
Using block or code programming, Sparky introduces programming from elementary-age to adults.

That being said, there is opportunity to create educational resources to teach AI and Robotics to children aiming to be, as pointed by the project JetBot, "affordable, educational and fun" \cite{nanoJetBot:2019} as well as to have child-centred programming languages.






\begin{table}
  \begin{tabular}{ccc}
    \toprule
    Project & Established  & Cost\\
    \midrule
    JPL Open Source Rover \cite{OSR:2018} & April 2018  &  USD 2500.00 \\
    JetBot AI Robot \cite{nanoJetBot:2019} & March 2019  & EUROS 212.00    \\
    Otto DIY robots \cite{OttoDIY:2016} & 2016 &  EUROS 100.00  \\
    Robot at AIR4Children & 2021 & EUROS 100.00  \\
  \bottomrule
\end{tabular}
\caption{Open source projects for educational AI and Robotics}
\label{tab:opensourceprojects}
\end{table}

\subsection{Open source hardware and software}
Adopting the philosophy of open source, AIR4Children is aiming to tackle the need of accessible and affordable resources for AI and Robotics to young audiences \cite{UNICEF2020}.
For instance, as a way to mitigate the hight prices of educational robots, our initial prototypes of AIR4Children are in the range of 100.00 EUROS. 
Such prototype, based on raspberry pi, arduino uno board and few actuators, is able to recognise voice commands in English language to move the robot in different directions (Fig \ref{fig:teaser}(a)).
Similarly, we have identified Otto robot, a DIY educational robot, which is based on arduino, servomotors and a scratch as interface to program the robot with various routines for sensors and actuators with a price of EURO 100.00 (Fig \ref{fig:tm} (b)). 
In addition to the release of JPL Open Source Rover by engineers at NASA in April 2018 with a price of 2500 USD \cite{OSR:2018} or
nano JetBot platform developed by engineers at NVIDIA in March 2019 with a price of 212 EUROS \cite{nanoJetBot:2019}. 
See Table~\ref{tab:opensourceprojects} that summarises open source projects with year of establishment and cost.

\section{AIR4Children as a open source project}
Having known the benefits of not only the lower price of open source projects but the increase of customisation and control of these, AIR4Children is therefore intending to adopt a similar journey along the lines of open source principles with the aim of making affordable, customisable and accessible tools for a child-centred AIR.

In a first phase, AIR4Children project will be piloting teaching materials with our open source robots and Otto, a well known open source educational robot \cite{OttoDIY:2016}.  
Then in a second phase, and with feedback of the pilots, teaching materials with child-centre programming languages as well as the customisation of open source robots will be improved to polish a more child-centred curriculum of AIR. 
On a third phase, children and adolescents in a range of age between 6 to 14 years old will be invited to enroll on workshops to be free of charge to all the participants. 
In this regard, this initial phases of the project will help us to provide evidence of the impact of AIR in a children of different backgrounds and evaluate children's perception on the fields of AI and Robotics.  

\subsection{Initial pilots of AIR4Children}
In 2013, a custom-make robot of 50USD were build to teach robotics to one child in which the challenging were to teach complex concepts in a fun way without any curriculum nor teaching materials. 
Then, in 2014, we managed to build a new robot prototype that make use of voice commands for easy interaction with children (Figure \ref{fig:teaser}(a)). 
With such prototype, we engaged with four children of a age range between 9 to 11, who enjoyed the voice command interaction of trying to move the robot in five commands (move forward, move backward, move right, move left and stop) (top image of Figure \ref{fig:teaser}(c)).
Recently, during three months from October 2020, we managed to virtually pilot Otto robot to one child accompanied with his mother for the constructions and the teaching of basis of AIR (bottom image of Figure \ref{fig:teaser}(c)).
Such pilots help us to clarify and to distill the aims of AIR4Children as a way to go for another round of pilots which will include curriculums and teaching materials to help both children, instructors (and perhaps their parents as well).

\subsection{Open teaching materials}
Considering teaching materials of AI and Robotics to be child-like oriented, AIR4Children is then adopting Montessori's education with the philosophy orbited around the quote “the hand is the instrument of the mind.” \cite{montessori2013absorbent}.
In that way, materials for AIR4Children are based on Montessori education to help children to internalise new concepts and to develop concentration of their learning skills through activities of movement and repetition.
One potential way to develop such skills is by designing activities in AI and robotics that are appropriately introduced in development stages \cite{bers2008, bers-horn2010, kazakoff-bers2012} as well as the development of grasping and understanding mathematical concepts (e.g. numbers, size, and shapes) and computational thinking (e.g. variables, loops, counters, etc) \cite{bers2012, resnick1998}.
Similarly, as Elkin et al. (2014) \cite{elkin2014} explained, AIR4Children can provide a way to engage children in problem-solving activities to allow children to participate in creative explorations, develop fine motor skills, hand-eye coordination, engage in collaborative and teamwork activities.

\begin{figure}[h]
  \centering
  \includegraphics[width=\linewidth]{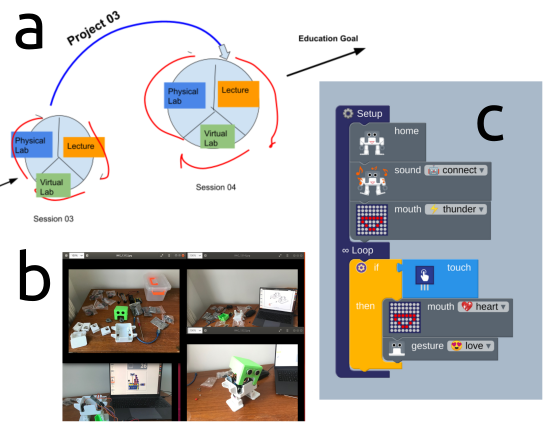}
  \caption{(a) Spiral learning divided into lecture, virtual and physical laboratory and connected via project to next session \cite{tarik2017} (b) robot assembly and (c) block-like programming style.}
  \label{fig:tm}
\end{figure}

That said, figure \ref{fig:tm}(a) illustrates the spiral learning technique adopted for AIR4Children to reinforce the above Montessori skills \cite{tarik2017}, fig \ref{fig:tm}(b) shows material of otto robot to develop creative exploration and engage in collaborative activities and fig \ref{fig:tm}(c) a block diagram as a child-centre programming language. 

\section{Conclusions and future work}
In this work, we coin the term AIR4Children as a project that aims to tackle aspects of inclusion, accessibility, transparency, equity, fairness and participation of children in the fields of AI and Robotics as well as to create teaching materials with a more child-centre approach for AI and Robotics.
We also touched on the open source projects in AI and Robotics as a corner stone for AIR4Children with the aim of minimise cost of the materials and made customised educational materials.  
We added our previous experience of few pilots since 2013 as way to narrow down the aims of AIR4Children.
Similarly, it has been presented the initial phases of AIR4Children that include piloting, implementing and refining workshops for children in the age range between 6 to 14 (to be implemented in Xicohtzinco, a town from Tlaxcala, M\'exico).
We also touched on the creation of curriculums following the philosophy of Montessori education, a non-traditional educational approach to help young audiences to develop skills to think creatively, with curiosity and open-minded and to develop a sense of wonder and joy for learning.

As a future work, AIR4Children aims to run workshops by the end of 2021 to put in practice educational material of child-centre AI and Robotics and to evaluate the impact on these fields to children, instructors, parents and their community. 


\begin{acks}
To Rocio Montenegro for her contributions and input as Montessori teacher. 
To Elva Corona for her contributions on giving vision to the goals of the project. 
To Marta P\'erez, Donato Badillo-Per\'ez and Antonio Badillo-Per\'e\ for their interest as initial testers of the project. 
To Leticia V\'azquez for her inputs as a teacher of teenagers. 
To Angel Mandujano for his help on the preparation of the software frameworks for the project. 
To Dago Cruz for his contributions on open source AI and Robotics.
To all of you of whom with the limited time were able to contribute to this work to make a bit more solid.
Thanks to Miguel Xochicale to orchestrate AIR4Children.
\end{acks}

\bibliographystyle{ACM-Reference-Format}


\end{document}